\documentclass[10pt,journal,twoside]{IEEEtran}

\usepackage{cite}

\ifCLASSINFOpdf
\else
\fi
\usepackage[ruled,vlined,linesnumbered]{algorithm2e}
\usepackage{amsmath}
\usepackage{amssymb}

\usepackage{array}

\usepackage{stfloats}
\usepackage{url}
\usepackage{hyperref}

\usepackage{xcolor}
\usepackage[nolist]{acronym}
\usepackage{graphicx}

\usepackage{makecell}
\usepackage{pdflscape}
\usepackage{adjustbox}
\usepackage{enumitem}
\usepackage{multirow}
\usepackage{pifont}
\usepackage{tabularx}
\usepackage{tablefootnote}
\usepackage{booktabs}
\usepackage{makecell}

\newcommand{\xmark}{\ding{55}}

\begin{document}

\begin{acronym}
    \acro{kyc}[KYC]{Know Your Customer}
    \acro{id}[ID]{Identity Document}
    \acro{ai}[AI]{Artificial Intelligence}
    \acro{ovd}[OVD]{Optical Variable Device}
    \acro{ocr}[OCR]{Optical Character Recognition}
    \acro{pai}[PAI]{Presentation Attack Instrument}
    \acro{pii}[PII]{Personally Identifiable Information}
    \acro{bid}[BID]{Brazilian Identity Document}
    \acro{pvc}[PVC]{Polyvinyl Chloride}

    \acro{pad}[PAD]{Presentation Attack Detection}
    \acro{ifdl}[IFDL]{Image Forgery Detection and Localization}
    
    \acro{cnn}[CNN]{Convolutional Neural Network}
    \acro{gan}[GAN]{Generative Adversarial Network}
    \acro{lora}[LoRA]{Low-Rank Adaptation}
    \acro{uld}[ULD]{Unconditional Latent Diffusion}
    \acro{ldm}[LDM]{Latent Diffusion Model}
    \acro{sam2}[SAM-2]{Segment Anything Model 2}
    \acro{eer}[EER]{Equal Error Rate}

    \acro{gimp}[GIMP]{GNU Image Manipulation Program}
    \acro{roi}[ROI]{Region Of Interest}
    
\end{acronym}

\title{FakeIDet3-DB: Refining Digital Attacks and Patch Extraction for Secure ID Benchmarking}

%
%
%

\author{Javier~Muñoz-Haro,
        Andres Teruel,
        Ruben Tolosana,
        Daniel DeAlcala,\\
        Ruben Vera-Rodriguez,
        Aythami Morales,
        Julian Fierrez
\thanks{All authors are with School of Engineering, Universidad Autónoma de Madrid, Ciudad Universitaria de Cantoblanco, Madrid, Spain.}
\thanks{Corresponding author: Javier Muñoz-Haro (javier.munnoz@uam.es).}%
}

%
%

\markboth{}
{Muñoz-Haro \MakeLowercase{\textit{et al.}}: Digital Attacks on IDs}
%



\maketitle

\begin{abstract}
Identity document (ID) authentication relies on the structural integrity of complex, high-frequency security patterns. However, advanced Generative AI models can now inject localized, high-fidelity manipulations, creating deceptive attacks that bypass standard verification. Training robust image forensic models to detect these anomalies is hindered by privacy regulations, forcing reliance on synthetic templates lacking the intricate visual patterns of real IDs. To bridge this domain gap, we introduce FakeIDet3-DB, the first comprehensive database of digital manipulations on real, government-issued IDs. FakeIDet3-DB encompasses classical (e.g., copy-move) and Generative AI-driven manipulations (e.g., face-swapping, inpainting) enhanced with advanced image refinement procedures to suppress visual artifacts. In addition, to comply with strict data protection regulations (e.g., GDPR), we adopt a recently-proposed framework based on patches. In order to maximize forensic utility, we formulate privacy-aware patch extraction from a real ID as a geometrically constrained image processing problem. We propose PACE, a Pseudo-Anonymized Contextual patch Extraction algorithm, which leverages Integral Image mapping and distance-driven Non-Maximum Suppression (NMS). PACE efficiently contours anonymization masks that prevent Personally Identifiable Information (PII) leakage while maximizing semantic density in peri-censorship regions, yielding almost 5.2M patches extracted from more than 6.4K images from real/fake IDs. Furthermore, an extensive evaluation of the proposed FakeIDet3-DB is performed using state-of-the-art models, showcasing they all struggle to detect and locate attacks coming from generative and classic techniques (32.45\% EER in detection and 83.48\% AUC-ROC in localization). Finally, we establish a privacy-compliant benchmark for both forgery detection and localization, demonstrating the challenges introduced by patch-based processing and real-world ID attacks. FakeIDet3-DB and the corresponding benchmark code are avaible at \url{https://github.com/BiometricsAI/FakeIDet3-DB}.
\end{abstract}

\begin{IEEEkeywords}
FakeIDet3-DB, Benchmark, Digital Attacks, Generative AI, Fake Identity Document, Algorithm.
\end{IEEEkeywords}

%
\IEEEpeerreviewmaketitle

\section{Introduction}

\begin{figure*}[!t]
    \centering
    \includegraphics[width=\textwidth]{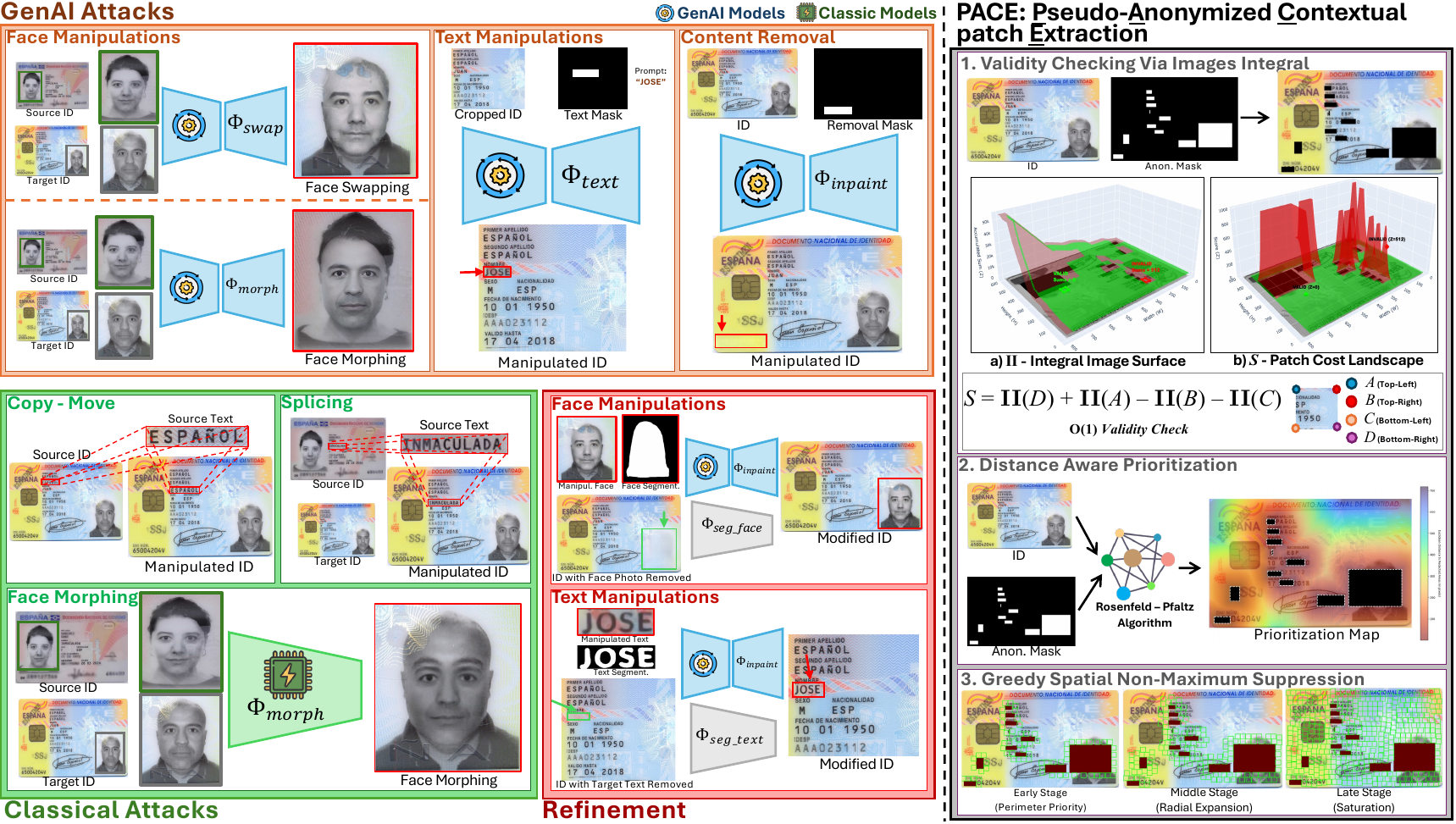}
    \caption{An overview of the main contributions of the present article. The left panel illustrates the attack generation pipeline considered in FakeIDet3-DB, integrating both classical methods and GenAI models which undergo a refinement post-processing to produce highly-realistic attacks. Right panel depicts the proposed PACE, an efficient, privacy-aware patch extraction algorithm that leverages Integral Images and a distance-aware greedy NMS to maximize residual sensitive information extraction around redacted sections without compromising any \ac{pii} from the \ac{id} owner.}
    \label{fig:graph_abs}
\end{figure*}

\IEEEPARstart{T}{he} proliferation of digital services has driven the transition from physical identity verification to remote, image-based authentication ecosystems. In standard \ac{kyc} protocols, users are required to submit smartphone-captured images of their physical identity document (ID). From an image processing perspective, authenticating these documents is a highly complex task, as the system must reliably analyze intricate spatial structures---such as \acp{ovd} (e.g., holograms, precise laser engravings)---under unconstrained acquisition conditions. This vulnerability is frequently exploited through injection attacks, where malicious actors bypass the camera sensor to directly feed manipulated \ac{id} images into the verification pipeline.

Concurrently, the rapid evolution of Generative AI (GenAI) models has fundamentally altered the landscape of digital image manipulation \cite{pedrouzo_avatars}. Rather than simply synthesizing images from scratch, these AI-driven tools can perform targeted, high-fidelity semantic alterations within existing visual content. Applied to \ac{id} tampering, techniques such as face-swapping \cite{tolosana2020deepfakes}, morphing \cite{face_morph_sur}, or text inpainting \cite{liu-etal-2023-character} act as localized signal perturbations. To evade detection, these GenAI attacks are often coupled with sophisticated image refinement procedures---such as feathering or illumination correction---that suppress tampering artifacts \cite{deep_watermarking} and seamlessly align the forged regions with the pristine source distribution.

Developing robust image forensic models \cite{for_det_reason, dom_adapt_fake} to detect these subtle structural anomalies in \acp{id} is hindered by a critical data acquisition bottleneck. Due to strict privacy regulations, existing \ac{id} databases rely almost exclusively on synthetic, laboratory-created templates \cite{fantasyid}. These templates lack the rich, high-frequency security patterns with complex and delicate visual forms from legitimately manufactured government \acp{id} (e.g., fine-grained engravings). Consequently, forensic models suffer from a severe domain shift: they achieve high accuracy on in-domain synthetic distributions but fail drastically when attempting to generalize to the topological and structural complexity of real-world manipulations \cite{Korshunov_2025_ICCV, tapia2025second}.

To overcome these fundamental limitations, we introduce FakeIDet3-DB, the first database of digital manipulations applied to real, government-issued \acp{id}. Adopting a recent privacy-aware framework \cite{fakeidet, fakeidet2}, FakeIDet3-DB strictly complies with data protection regulations by applying irreversible modifications---pitch-black anonymization masks---and extracting patches from non-redacted areas, thus preventing any Personally Identifiable Information (PII) leakage. 

While anonymization solves the privacy bottleneck, it introduces severe visual artifacts that hinder standard feature extraction. To address this, we formulate patch extraction as a geometrically constrained image processing problem. We propose an optimized, context-aware extraction algorithm that utilizes Integral Image mapping ($\mathcal{O}(1)$ local complexity) \cite{kochkarev2020data} and distance-driven Non-Maximum Suppression (NMS). This algorithmic approach effectively contours the complex geometries of the anonymization masks, targeting the highly informative peri-censorship regions where residual sensitive data and manipulation artifacts typically reside, while strictly guaranteeing zero censorship contamination. 

The main contributions of this article are summarized in Fig.~\ref{fig:graph_abs} and as follows:

\begin{itemize}
    \item \textbf{FakeIDet3-DB:} The most comprehensive \ac{id} database to date for digital manipulations. Sourced from 250 real images of 47 distinct government-issued \acp{id}, we generated 8 different attack typologies, resulting in over 6.4K images of real/fake \acp{id}. These range from \textit{Classical Manipulations} (e.g., copy-move, splicing) to \textit{GenAI Manipulations} (e.g., face-swapping, morphing, or text inpainting). Crucially, these attacks are enhanced with advanced image refinement procedures to minimize visual artifacts and simulate high-quality, real-world forgeries. 
    \item \textbf{Privacy-Aware Patch Extraction Algorithm:} We introduce a \underline{P}seudo-\underline{A}nonymized \underline{C}ontextual patch \underline{E}xtraction algorithm (PACE). By leveraging global spatial optimization rather than naive uniform grids, our algorithm significantly increases the semantic density of the extracted dataset, maximizing the retention of semantically-relevant artifacts located at the boundaries of occluded regions. As a result, we extract over 5.2M patches from pseudo-anonymized \ac{id} images at both 64$\times$64 and 128$\times$128 patch sizes, following our previous framework \cite{fakeidet2}.
    \item \textbf{Standardized Benchmark:} To foster this challenging line of research within the image processing and forensics community, we design and release a reproducible benchmark\footnote{https://github.com/BiometricsAI/FakeIDet3-DB} for both the detection and spatial localization of fake \acp{id}, providing access to real-world document evaluations without compromising \ac{pii}.
\end{itemize}

The remainder of this article is structured as follows. Section \ref{sec:rel_works} reviews current \ac{id} databases, competitions and patch extraction methodologies, addressing their primary limitations. Section \ref{sec:fakeidet3db} introduces FakeIDet3-DB, detailing the image manipulation and refinement processes. Section \ref{sec:patch_extraction} formulates the proposed context-aware patch extraction algorithm under spatial constraints. In Section \ref{sec:eval_bchmrk}, we evaluate state-of-the-art media forensics architectures against FakeIDet3-DB, and finally, Section \ref{sec:conclusion} draws the conclusions of this work.

\section{Related Work}
\label{sec:rel_works}

\subsection{Fake ID Databases and Competitions}
\label{sub:databases_comps}
Table \ref{tab:db_summary} depicts the landscape of fake \ac{id} detection databases, which has historically relied on proxy representations due to strict privacy constraints regarding real \acp{id}. Early databases focused on physical forgeries, utilizing either laboratory-manufactured replicas, such as the KID34K dataset \cite{kid34k}, or digital templates printed on \ac{pvc}, like the MIDV database family and DLC-2021 \cite{bulatov2020midv, Bulatov2022, arlazarov2019midv, polevoy2022document}. More recently, addressing the rise of digital tampering, several works have introduced entirely synthetic databases (e.g., IDNet \cite{idnet}, providing over 800K generated samples, and DASAC \cite{dasac}) or digitally manipulated fictitious templates (e.g., FantasyID \cite{fantasyid}). While these databases offer large-scale volume and diverse attack typologies, they inherently lack the rich, high-frequency security features of real, government-issued \acp{id}. To tackle the sensitive nature of real \acp{id}, FakeIDet2-db \cite{fakeidet2} proposed a patch-based framework utilizing pitch-black redacted areas over \ac{id} owners' sensitive data. However, its scope was exclusively limited to physical forgeries, leaving challenging digital manipulations on real \acp{id} unexplored.

The fundamental limitation of relying on synthetic or lab-created proxies as ``real" \acp{id} is the severe domain shift they introduce. Recent international evaluations have severely exposed this vulnerability. In the Second PAD-ID Competition \cite{tapia2025second}, while the winning solution achieved a 6.36\% \ac{eer} by leveraging a massive proprietary database, runner-up models evaluated against government-issued real \acp{id} suffered severe performance degradation (\acp{eer} of 23.87\% and 31.94\%) compared to evaluations on lab-made proxies. Similarly, in the DeepID challenge \cite{Korshunov_2025_ICCV}, forensic models trained on synthetic data achieved a near-perfect F1 score of 0.99 on synthetic test sets, but their performance plummeted to 0.72 when evaluated on government-issued real \acp{id} provided by industry partners (PXL Vision). This substantial performance degradation underscores the urgent need for public, open-source databases featuring government-issued real \acp{id} to democratize research and bridge the gap between laboratory proxies and real fraud.

\begin{table*}
\hspace*{-0.2cm}
\centering
    \caption{Main characteristics of public fake ID databases in the literature. We point out whether the databases contain physical or digital attacks. Refined Attacks refer to databases containing attacks that undergone a post-processing or fine-tuning procedure to create high-quality manipulations.}
    \begin{tabular}{cccccccccc}
        \toprule
        \textbf{Database} & \textbf{\#Samples} & \textbf{\#IDs} & \textbf{\makecell{\#Devices}} & \textbf{Attack Types} & \textbf{\#Attacks} & \textbf{\makecell{Government-Issued \\ Real IDs?}} & \textbf{\makecell{Refined \\ Attacks?}} & \textbf{Public?} \\
        \midrule
        \makecell[c]{DLC-2021 (2021) \\ \cite{polevoy2022document}} & 1,424 (videos) & 1,000 & 2 & Physical & 4 & \textcolor{red}{\xmark} & N/A &\textcolor{green}{\checkmark} \\
        \makecell[c]{Benalcazar \textit{et al.} (2023) \\ \cite{synth_id_card_db}} & 3,000 & 3,000 & N/A & Digital & 1 & \textcolor{red}{\xmark} & \textcolor{red}{\xmark} & \textcolor{green} {\checkmark} &  \\
        \makecell[c]{KID34K (2023) \\ \cite{kid34k}} & 34,662 & 92 & 12 & Physical & 3 & \textcolor{red}{\xmark} & N/A & \textcolor{green}{\checkmark} \\
        \makecell[c]{IDNet (2024) \\ \cite{idnet}} & 837,060 & 837,060 & Synthetic & Digital & 7 & \textcolor{red}{\xmark} & \textcolor{red}{\xmark} & \textcolor{green}{\checkmark} \\
        \makecell[c]{FantasyID (2025) \\ \cite{fantasyid}} & 3,284 & 362 & 3 & Digital & 3 & \textcolor{red}{\xmark} & \textcolor{green}{\checkmark} & \textcolor{green}{\checkmark} \\
        \makecell[c]{DASAC (2025) \\ \cite{dasac}} & 150,833 & 1,472 & 3 & Digital, Physical & 5 & \textcolor{red}{\xmark} & \textcolor{red}{\xmark} & \textcolor{green}{\checkmark} \\
        \makecell[c]{FakeIDet2-db (2026) \\ \cite{fakeidet2}} & 2,000 & 47 & 3 & Physical & 3 & \textcolor{green}{\checkmark} & N/A & \textcolor{green}{\checkmark} \\
        \hline
        \makecell[c]{FakeIDet3-DB \\ \textbf{(proposed)}} & 6,436 & 47 & 3 & Digital & 8 & \textcolor{green}{\checkmark} & \textcolor{green}{\checkmark} & \textcolor{green}{\checkmark} \\
        \bottomrule
    \end{tabular}
    \label{tab:db_summary}
\end{table*} 

\subsection{Geometrical Constrained Patch Extraction} 
Spatial patch extraction has evolved significantly beyond rigid uniform grids \cite{fakeidet2}. PatchMatch \cite{barnes2009patchmatch} pioneered stochastic perturbation for rapid spatial mapping. In medical and geospatial domains, Kochkarev \textit{et al.} \cite{kochkarev2020data} introduced probabilistic sampling via distance transforms and integral images to balance rare classes. Additionally, in-context learning frameworks like PatchICL \cite{ndir2026scaling} leverage boundary-guided Gumbel-Top-K sampling for selective processing. However, all these patch extraction methods lack the deterministic, zero-overlap topological constraints mandated by privacy and data regulation laws, which is a critical gap addressed by our proposed distance-driven patch extraction framework.

\begin{figure*}[!t]
    \centering
    \includegraphics[width=\textwidth]{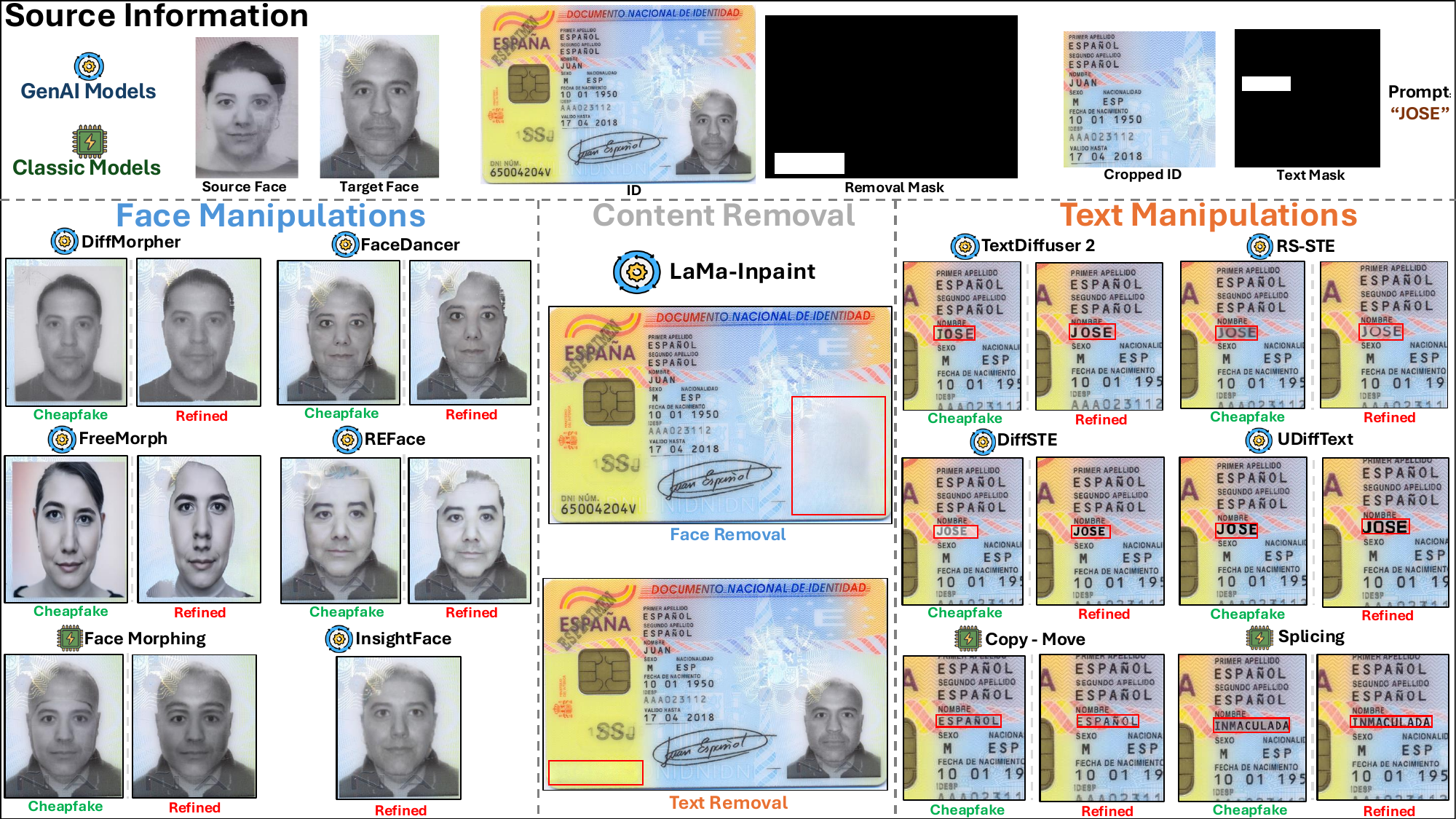}
    \caption{Visual results of the digital attacks included in FakeIDet3-DB including different types of manipulations and post-processing techniques. It is noticeable how refined attacks remove clear artifacts left by text inpainting models, seamlessly blending background with foreground text. For face manipulations, we observe that for GenAI face morphing attacks, our refined pipeline contours the morphed subject and faithfully blends it with the background canvas. Note that for content-removal attacks, no post-processing is needed.}
    \label{fig:attacks_qa}
\end{figure*}

\section{FakeIDet3-DB: Attacks Generation}
\label{sec:fakeidet3db}

As demonstrated in previous section, the persistent lack of government-issued  real \acp{id} in research databases introduces a severe domain gap that limits real-world applicability. To overcome this fundamental barrier, we introduce the FakeIDet3-DB database. By leveraging proven privacy-aware frameworks \cite{fakeidet, fakeidet2}, this database securely utilizes real \ac{id} data without compromising \ac{pii}.

FakeIDet3-DB includes two main types of digital attacks. \textit{Classical Attacks} use existing content and classical algorithms (e.g., copy-move, splicing, and landmark-based face morphing). Conversely, \textit{GenAI Attacks} leverage generative models to alter real content. Both types undergo a refined post-processing pipeline to faithfully embed the modified elements, producing attacks with varying degrees of visual realism. Fig. \ref{fig:attacks_qa} shows attacks generated under identical conditions with and without our proposed refinement pipeline (i.e., \textit{refined} vs. \textit{cheapfake}).

\subsection{Digital Attacks Formulation}
\label{ssec:digital_attacks}

To generate the digital attacks, we utilize the 250 real \acp{id} introduced in FakeIDet2-db \cite{fakeidet2}. The synthesis is modeled as a set of image-domain transformations, categorized into classical signal-processing attacks and GenAI-driven manipulations.

\subsubsection{\textbf{Classical Attacks}}
\label{sssec:classical_attacks}
Classical attacks rely on the spatial redistribution of existing pristine pixels. These manipulations are divided into text-based and face-based transformations.

For text-based manipulations, the methodology consists of a spatial mapping of a source text region $\mathcal{B}_{src}$ from a source image $I_{src}$ into a target bounding box $\mathcal{B}_{tgt}$ in a target image $I_{tgt}$. To ensure the forgery remains visually pristine, semantic compatibility is strictly enforced (e.g., date fields are exclusively mapped to date fields). We formulate two widely recognized spatial transformations \cite{img_for}:
\begin{itemize}
    \item \textbf{Copy-Move (Intra-document):} The manipulation is constrained such that $I_{src} = I_{tgt}$. Furthermore, fine-grained attacks are generated by restricting $\mathcal{B}_{src}$ to sub-character topological regions (e.g., duplicating a single digit), introducing high-precision spatial alterations.
    \item \textbf{Splicing (Inter-document):} The manipulation maps content between disjoint image domains, where $I_{src} \neq I_{tgt}$.
\end{itemize}

Face-based manipulations encompass portrait swapping and face morphing. \textit{Portrait swapping} functions as a facial splicing attack (inter-document mapping of the portrait region), ensuring gender-matched identities. For \textit{face morphing}, we utilize open-source frameworks\footnote{\url{https://github.com/alyssaq/face_morpher}}\textsuperscript{,}\footnote{\url{https://github.com/Azmarie/Face-Morphing/}} based on classical geometric warping. Given a source face, denoted as $F_{src}$ and a target face, denoted as $F_{tgt}$, facial landmarks $L_{src}$ and $L_{tgt}$ are extracted. Delaunay triangulation \cite{face_morph_sur} is used to compute a spatial affine warping function $\mathcal{W}$, aligning both faces to a mean geometry. The morphed face $F_{morph}$ is then synthesized via linear alpha blending:
\begin{equation}
    F_{morph} = \lambda \cdot \mathcal{W}(F_{src}, L_{src}, L_{tgt}) + (1-\lambda) \cdot \mathcal{W}(F_{tgt}, L_{tgt}, L_{src})
\end{equation}
where $\lambda \in [0, 1]$ controls the blending factor.

\begin{figure*}[!t]
    \centering
    \setlength{\fboxsep}{0pt}
    \fbox{\includegraphics[width=0.89\textwidth]{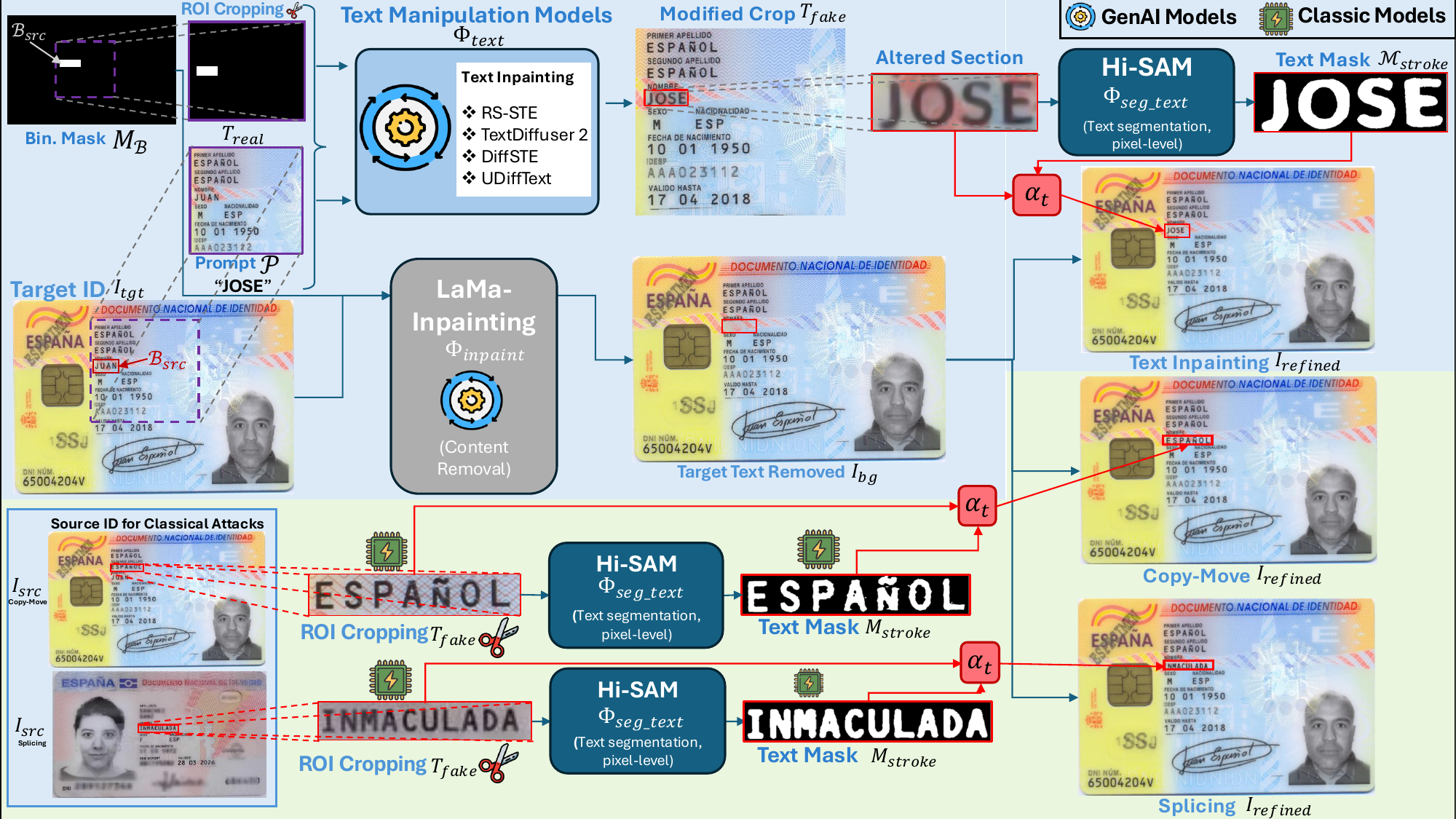}}
    \caption{Proposed refined post-processing pipeline for text digital attacks. The GenAI pipeline (top, light-blue background) involves GenAI methods ($\Phi_{text}$) to synthesize the new text, while also removes the target text area ($\Phi_{inpaint}$), creating $I_{bg}$. Finally, an alpha-blending operation $\alpha_{t}$ from the extracted text mask is applied to embed the new content. The classical pipeline (bottom, light-green background) leverages the existing content ($I_{src}$) along with the target text removed image ($I_{bg}$) and embeds the source content crop $\mathcal{T}_{fake}$ into the $I_{tgt}$ in a similar fashion as the GenAI pipeline (i.e., alpha-blending operation).}
    \label{fig:refined_text_attacks}
\end{figure*}

\subsubsection{\textbf{GenAI Attacks}}
\label{sssec:genai_attacks}

In generative AI-driven digital attacks, the forgery process is conceptualized as a conditional generative function. To ensure high-quality synthesis, we employ \acp{ldm} \cite{diffmodels_surv}  for text inpainting (DiffSTE \cite{diffste}, UDiffText \cite{udifftext}, TextDiffuser 2 \cite{textdiff2}, RS-STE \cite{rsste}), face morphing (DiffMorpher \cite{diffmorph}, FreeMorph \cite{freemorph}) and face swapping (REFace \cite{reface}). We also rely on \acp{gan} \cite{gan_rev} for face swapping (FaceDancer \cite{facedancer}, InsightFace\footnotemark) and content removal (LaMa-Inpaint \cite{lama_inpaint}). This multi-model strategy endows FakeIDet3-DB with a wide spectrum of visual manipulation qualities and distinct generation fingerprints \cite{wang2020cnn, sinitsa2024deep, ganprintr}, forcing detection models to learn robust, architecture-agnostic features. 

\footnotetext{\href{https://github.com/deepinsight/insightface}{https://github.com/deepinsight/insightface}}

For text-based manipulations, the generative process is modeled as an inpainting function $\Phi_{text}$. We select a target bounding box $\mathcal{B}_{tgt}$ and create a binary mask $M_\mathcal{B}$ along with a contextual text prompt $\mathcal{P}$. Crucially, we account for the structural limitations of current text-inpainting LDMs, which struggle with character-space mismatches. Therefore, we impose a strict character-length constraint between the original legitimate string and the synthesized prompt: $|\mathcal{P}| = |\text{original text}|$. For names and surnames, we sample replacements from a Hispanic demographic database. For the \ac{id} number, valid modulo-23 sequences are algorithmically generated. To provide sufficient receptive field for font and texture replication, an expanded center crop $T_{tgt}\subset I_{tgt}$ (e.g., $768 \times 768$) is processed:
\begin{equation}
    T_{fake} = \Phi_{text}(T_{tgt}, \mathcal{P}, M_{\mathcal{B}})
\end{equation}
The newly generated text crop $T_{fake}$ is subsequently extracted for downstream integration.

For face manipulations, portrait crops are either automatically or manually extracted from the target $I_{tgt}$ and source $I_{src}$ images, yielding $F_{src} \subset I_{src}$ and $F_{tgt} \subset I_{tgt}$. The generative model $\Phi_{face}$ maps the source and target characteristics with the operation $F_{fake} = \Phi_{face}(F_{src}, F_{tgt})$ to generate a realistic synthetic face $F_{fake}$, which is reserved for subsequent seamless blending.

Furthermore, we introduce a Content Removal manipulation, which acts as a purely subtractive alteration. Using the dataset's metadata, a random field is targeted, and its spatial binary mask $M_{\mathcal{B}}$ is generated. Using an inpainting network $\Phi_{inpaint}$ (e.g., LaMa-Inpaint \cite{lama_inpaint}), we apply the operation $I_{removed} = \Phi_{inpaint}(I_{tgt}, M_{\mathcal{B}})$, which removes the foreground semantic information while explicitly emulating the underlying high-frequency security patterns. 

In both text and face manipulations, synthesizing the raw content ($T_{fake}$ or $F_{fake}$) is only the first step. The realistic embedding of these forgered pixels into the complex visual domain of a real \ac{id} document is critical to create a faithful attack. The following section details the post-processing pipelines designed to achieve this seamless integration.

\begin{figure*}[!t]
    \centering
    \includegraphics[width=0.89\textwidth]{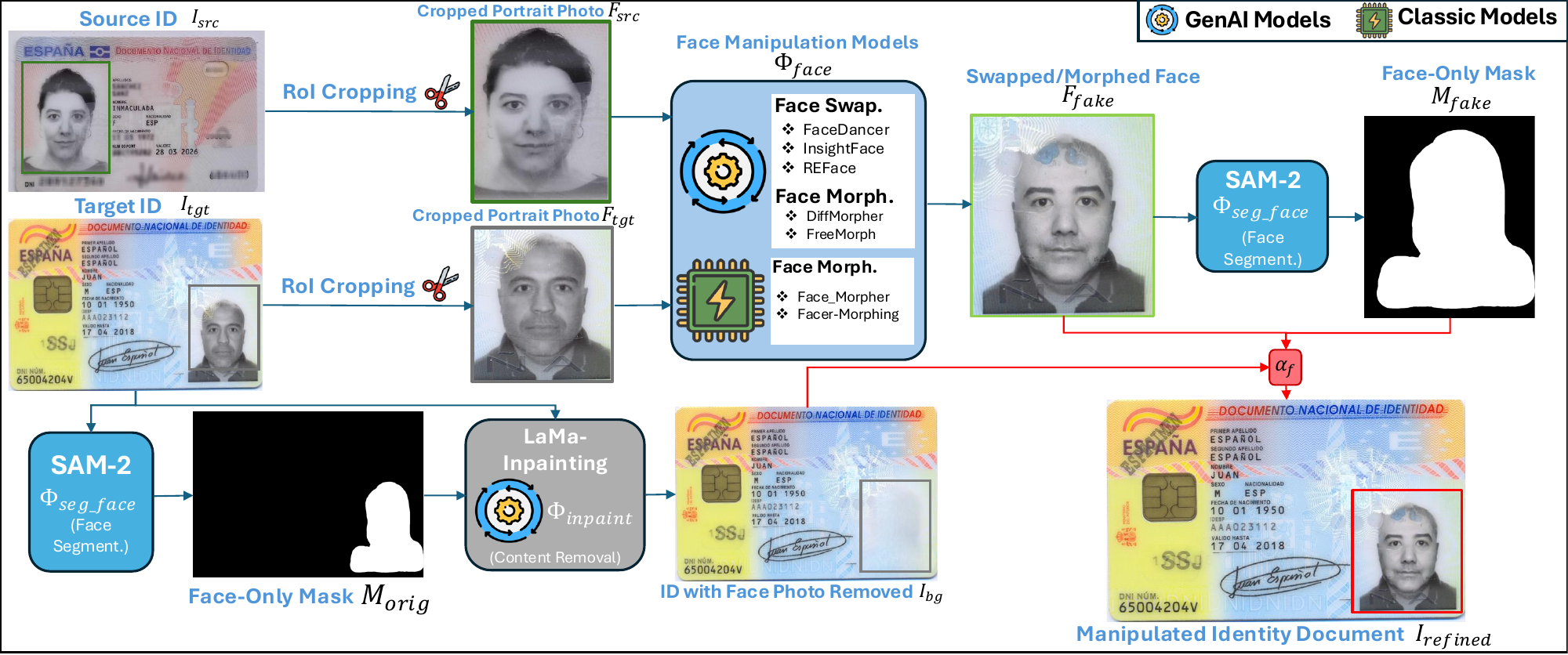}
    \caption{Proposed refined post-processing pipeline for face digital attacks. Given two images, source ($I_{src}$) and target ($I_{tgt}$), a portrait photo crop is applied to both images to restrict the context to the \ac{id} owner's face. Both images are then forwarded into a generic face manipulation function $\Phi_{face}$, which can either be classical algorithms or GenAI models. Once the manipulated $F_{fake}$ image is generated, a binary mask locating the areas of the new face is produced by $\Phi_{seg\_face}$. In parallel, the \ac{id} owner face is segmented, obtaining a mask that conditions the $\Phi_{inpaint}$ function to remove the portrait photo area, yielding $I_{bg}$. Finally, an alpha-blending operation $\alpha_{f}$ embeds the new manipulated face in $I_{bg}$, creating the final $I_{refined}$.}
    \label{fig:ref_face_attacks}
\end{figure*}

\subsection{Post-Processing Refinement Pipeline}
\label{ssec:postprocessing_pipeline}

We describe next how the manipulated content generated in previous section is seamlessly embedded into the digital attacks. In this work, \textit{cheapfake} refers to low-effort forgeries through a naive spatial substitution strategy between target ($I_{tgt}$) and source ($I_{src}$). The source content (coming from a generative model or existing content) is scaled and placed directly over the target using a static bounding region via hard-pixel replacement without any post-processing. Because this generic masking forces the source content into a rigid spatial constraint rather than respecting the semantic contours, it introduces unnatural boundary transitions. In the following sections, we formalize the post-processing techniques employed to create manipulations with varying degrees of visual realism. The pipeline diverges slightly depending on the manipulation modality (text vs. face).

\subsubsection{\textbf{Text-based Manipulations}}
\label{sssec:text_manipulations}

To overcome the visual discrepancies originated from cheapfake attacks, we introduce a \textit{refined} post-processing pipeline (Fig. \ref{fig:refined_text_attacks}) for high-fidelity manipulations. This framework models text integration as a semantic alpha-blending problem. First, to prevent structural artifacts from the original text, the pipeline reconstructs the background canvas. Let $M_{\mathcal{B}} \in \{0,1\}^{H \times W}$ be the binary mask of the target field. We process $I_{tgt}$ and $M_{\mathcal{B}}$ through an inpainting network $\Phi_{inpaint}$ (e.g., LaMa-Inpaint \cite{lama_inpaint}) applying the operation $I_{bg} = \Phi_{inpaint}(I_{tgt}, M_{\mathcal{B}})$ to generate the underlying document patterns (e.g., security grids, fine-grained engravings, etc).

Concurrently, let $T_{fake}$ denote the text crop synthesized by the generative model. To isolate the typographic ink from any mismatched generated background, we apply a high-fidelity text segmentation network $\Phi_{seg\_text}$ (e.g., Hi-SAM \cite{hisam}) operating in a sliding-window patch mode, $M_{stroke} = \Phi_{seg\_text}(T_{fake})$, where $M_{stroke} \in \{0,1\}^{H \times W}$ is a pixel-level text stroke mask.

Unlike hard-pixel masking, text strokes require subtle edge softening to blend naturally with the document's resolution. We apply a Gaussian smoothing kernel $G_{\sigma_{t}}$ (e.g., a $3 \times 3$ kernel) to generate a soft alpha matte $\alpha_{t} \in [0,1]^{H \times W}$:
\begin{equation}
    \alpha_{t} = M_{stroke} \ast G_{\sigma_{t}}
\end{equation}

The final composition is driven by the Hadamard product ($\odot$), transferring only the feathered typographic ink onto the clean background at the target \ac{roi}:
\begin{equation}
    I_{refined} = \alpha_{t} \odot T_{fake} + (\mathbf{1} - \alpha_{t}) \odot I_{bg}
    \label{eq:alpha_blend_text}
\end{equation}
This integration strictly preserves the authentic security features of $I_{bg}$, yielding sophisticated forgeries that significantly increase forgery detection and localization difficulty.

\subsubsection{\textbf{Face-based Manipulations}}
\label{sssec:face_manipulations}

To mitigate structural artifacts characteristic from cheapfake, we propose a semantic-aware \textit{refined} pipeline for face manipulations, depicted in Fig. \ref{fig:ref_face_attacks}. This framework abandons rigid bounding boxes in favor of dynamic semantic segmentation and spatial alpha matting. 

First, we isolate the original subject to reconstruct the background canvas. We obtain a precise mask of the real portrait photo area $M_{orig} \in \{0,1\}^{H \times W}$ using a promptable segmentation model $\Phi_{seg\_face}$ (e.g., SAM-2 \cite{sam2}). To guarantee the complete removal of the original facial contour (including hair) and prevent peripheral artifacts during background reconstruction, we apply a massive morphological dilation operation to $M_{orig}$ using a large structuring element $K$ (e.g., a $75 \times 75$ kernel), denoted as $M_{erase} = M_{orig} \oplus K$. The background canvas is then generated as $I_{bg} = \Phi_{inpaint}(I, M_{erase})$.

Concurrently, the pipeline semantically segments the generative or classic model's raw output $F_{fake}$ to extract the precise manipulated contours, discarding peripheral mismatched pixels: $M_{fake} = \Phi_{seg\_face}(F_{fake})$. To ensure a seamless transition between the synthesized skin/hair and the document background, $M_{fake}$ is softened into an alpha matte $\alpha_{f} \in [0,1]^{H \times W}$ via a Gaussian kernel $G_{\sigma_{f}}$ (e.g., a $5 \times 5$ kernel):
\begin{equation}
    \alpha_{f} = M_{fake} \ast G_{\sigma_{f}}
\end{equation}

The final content-aware composition is achieved via alpha blending:
\begin{equation}
    I_{refined} = \alpha_{f} \odot F_{fake} + (\mathbf{1} - \alpha_{f}) \odot I_{bg}
    \label{eq:alpha_blending_face}
\end{equation}
By replacing rigid templates with semantic isolation and background inpainting, this formulation effectively eliminates sharp boundaries, yielding highly realistic visual forgeries.

Moreover, we subject a single authentic \ac{id} to multiple simultaneous refined manipulations to simulate a comprehensive identity forgery scenario. In this context, an attacker compromises several fields concurrently---for instance, by replacing textual personal data in conjunction with face morphing manipulations of the target subject. We define such forgeries as \textit{multi-attacks}.

\section{FakeIDet3-DB: Proposed Patch Extraction Methodology}
\label{sec:patch_extraction}

\begin{algorithm}[!t] 
\caption{Proposed PACE Patch Extraction}
\label{alg:patch_extraction}

\KwIn{Image $\mathbf{I} \in \mathbb{R}^{H \times W \times C}$, Anon. Mask $\mathbf{M}_A \in \{0, 1\}^{H \times W}$, GT Mask $\mathbf{M}_{GT} \in \{0, 1\}^{H \times W}$, Patch Size $(h, w)$, Evaluation Stride $s$}
\KwOut{Extracted Patches $\mathcal{E}_{\mathbf{I}}$, Extracted Ground Truth Masks $\mathcal{E}_{GT}$}

\BlankLine
$\mathcal{E}_{\mathbf{I}} \gets \emptyset, \quad \mathcal{E}_{GT} \gets \emptyset$, \quad $\mathcal{C} \gets \emptyset$, \quad $\mathcal{V} \gets \emptyset$ 

\BlankLine
\tcp{1. $\mathcal{O}(1)$ Validity Checking via Integral Images}
$\mathbf{II}(x,y) \gets \sum_{x' \le x, y' \le y} \mathbf{M}_A(x', y')$ \;

\For{$(x, y) \in [0, W-w] \times [0, H-h]$ \KwTo \text{with step } $s$}{
    $S \gets \mathbf{II}(x+w, y+h) - \mathbf{II}(x, y+h) - \mathbf{II}(x+w, y) + \mathbf{II}(x, y)$\;
    \If{$S = 0$}{
        $\mathcal{V} \gets \mathcal{V} \cup \{(x,y)\}$ \tcp*{Mark as valid coordinate}  
    }
}

\BlankLine
\tcp{2. Distance-Aware Prioritization}
$\mathbf{D} \gets \text{DistanceTransform}(1 - \mathbf{M}_A)$\;
$\mathcal{C} \gets \text{Sort}(\mathcal{V}, \text{key} \gets \mathbf{D}(x + \frac{w}{2}, y + \frac{h}{2}), \text{order} \gets \text{ascending})$\;

\BlankLine
\tcp{3. Greedy Spatial Non-Maximum Suppression (NMS)}
$\mathbf{O} \gets \mathbf{0}^{H \times W}$

\For{$(x,y) \in \mathcal{C}$}{
    \If{$\sum \mathbf{O}[y:y+h, x:x+w] = 0$}{
        $\mathbf{O}[y:y+h, x:x+w] \gets 1$ \tcp*{Mark area as occupied}
        $\mathcal{E}_{\mathbf{I}} \gets \mathcal{E}_{\mathbf{I}} \cup \{\mathbf{I}[y:y+h, x:x+w]\}$\;
        $\mathcal{E}_{GT} \gets \mathcal{E}_{GT} \cup \{\mathbf{M}_{GT}[y:y+h, x:x+w]\}$\;
    }
}
\Return $\mathcal{E}_{\mathbf{I}}, \mathcal{E}_{GT}$\;
\end{algorithm}

Sharing complete, uncensored real \acp{id} introduces severe privacy and regulatory risks. To mitigate this vulnerability, patch-based processing \cite{patch_face_forgery, patchforensics} enables a secure collaboration paradigm between data-rich ID Holders and technology-rich AI Researchers as introduced in \cite{fakeidet2}. Under this privacy-aware framework, ID Holders locally censor sensitive information, extract valid patches, and shuffle them to completely dismantle the document's spatial layout. Furthermore, \ac{id} censoring can be applied at different levels. As per \cite{fakeidet2}, sensitive data can be completely occluded (i.e., fully-anonymized \ac{id}) or partially occluded (i.e,. pseudo-anonymized \ac{id}) leaving small, residual sections of the sensitive data uncovered which do not compromise the owners' \ac{pii}. Consequently, AI models can be safely trained (e.g., through self-supervised learning) on these strictly anonymized patches before being deployed back to the ID Holders for secure, local inference.

\begin{figure}
    \centering
    \includegraphics[width=0.7\columnwidth]{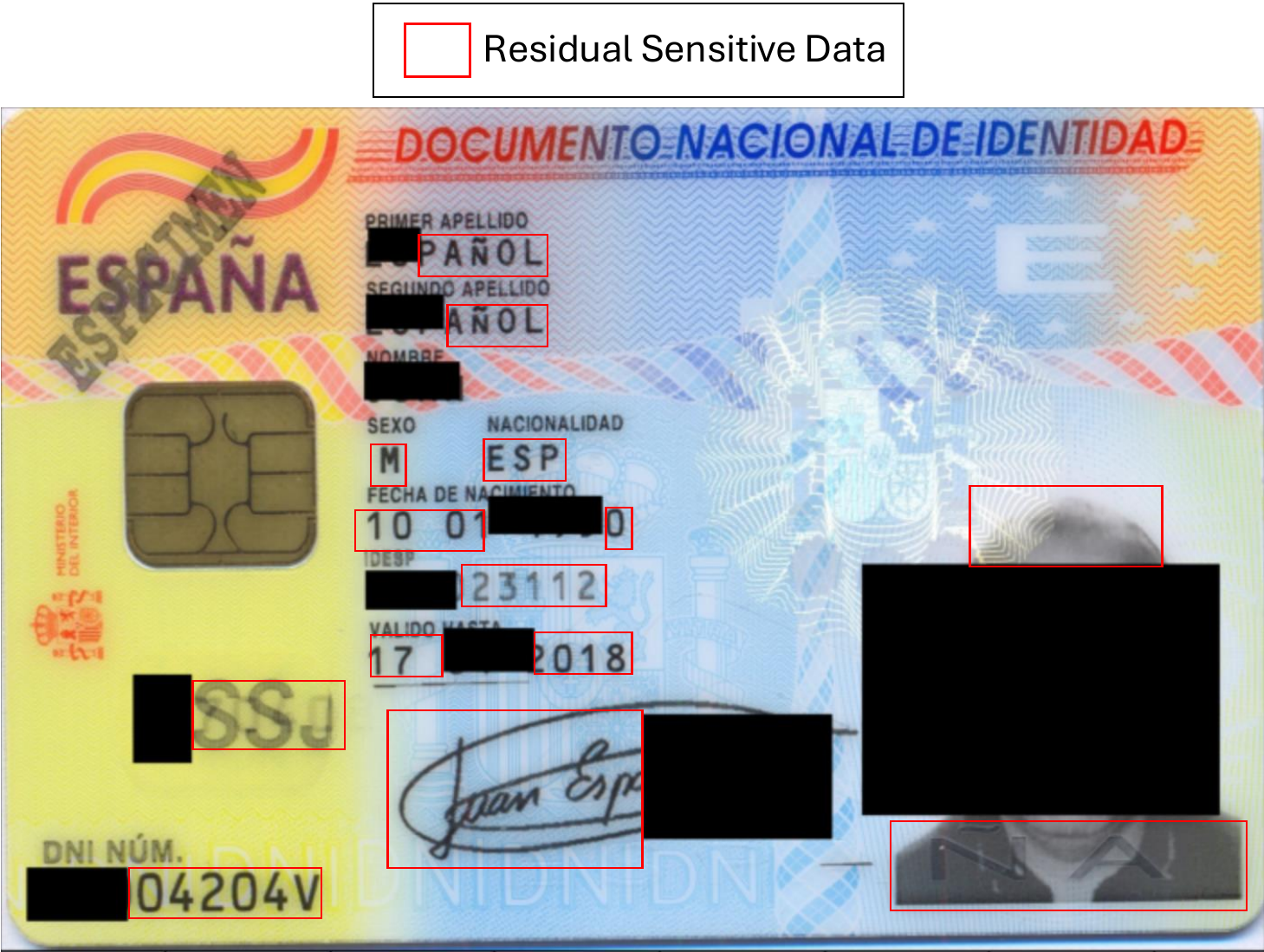}
    \caption{Visual example of a \textit{pseudo-anonymized ID}, where redacted sections make practically impossible to infer the sensitive information from the owner, thus preventing PII leaking. Uncovered sections are denoted as residual sensitive data, which are important for targeted patch extraction.}
    \label{fig:pseudo-anon}
\end{figure}

While this paradigm successfully protects privacy, anonymization artifacts (i.e., pitch-black rectangles) severely poison feature extraction. Naive uniform grid algorithms used in \cite{fakeidet2} fail to address this topological challenge, suffering from both semantic dilution (over-representing uninformative background) and censorship contamination (inadvertently accepting partially redacted cells). Because censored areas mimic digital manipulations, training on contaminated patches heavily degrades attack detection. Conversely, in pseudo-anonymized configurations \cite{fakeidet2}, the most discriminative forensic features---such as \textit{residual sensitive data} and local manipulation artifacts---reside strictly at the immediate boundaries of these occluded regions as shown in Fig. \ref{fig:pseudo-anon}. 

To overcome these limitations, we propose an optimized, context-aware patch extraction approach (Algorithm \ref{alg:patch_extraction}), summarized visually in Fig.~\ref{fig:graph_abs} (right panel). Our proposal considers a geometrically constrained greedy strategy designed to precisely target and maximize peri-censorship information while strictly guaranteeing zero overlap with the redacted pixels.

\subsection{Proposed Patch Extraction Method: PACE}
The enforced objective for the proposed PACE algorithm is a constrained spatial packing problem: we must extract the maximum number of valid patches that are as strictly close to the anonymization mask $\mathbf{M}_A$ as possible, without any overlap with the mask itself or between the patches. Naively evaluating every possible patch candidate involves nested loops with a computational complexity of $\mathcal{O}(H \cdot W \cdot h \cdot w)$, where $H$ and $W$ are the height and width of the image and $h$ and $w$ are the height and width of the patches. As this is intractable for large-scale datasets, we bypass this bottleneck via a three-stage optimization pipeline, described next.

\subsubsection{\texorpdfstring{$\mathcal{O}(1)$}{O(1)} Validity Checking via Integral Images}To instantaneously determine if a proposed patch contains any forbidden (anonymized) pixels, we compute the Summed-Area Table, or Integral Image $\mathbf{II}$ \cite{kochkarev2020data}, of the binary anonymization mask $\mathbf{M}_A$. The value at any location $(x,y)$ in $\mathbf{II}$ represents the sum of all pixels above and to the left of that coordinate.

Consequently, the sum of pixels $S$ (i.e., the cost) within any rectangular region bound by $(x, y)$ and $(x+w, y+h)$ can be computed in $\mathcal{O}(1)$ time using exactly four array references:
\begin{equation}
    S = \mathbf{II}(x+w, y+h) + \mathbf{II}(x, y) - \mathbf{II}(x, y+h) - \mathbf{II}(x+w, y) 
\end{equation}
If the cost is zero (i.e., $S = 0$), then the patch is considered strictly valid ($\mathcal{V}$). This reduces the search complexity for valid coordinates from $\mathcal{O}(h \cdot w)$ per patch to $\mathcal{O}(1)$. Furthermore, we introduce the stride parameter $s$, which allows to skip adjacent coordinates in both $x$ and $y$ axes. Since close coordinates contain highly correlated content, this reduces the number of valid coordinates by a factor of $s^2$. 

\subsubsection{Distance-Aware Prioritization}
Identifying valid patches is insufficient, as those physically adjacent to the occlusions (i.e., those containing residual sensitive information) must be prioritized. We apply a Distance Transform \cite{rosenfeld1966sequential, rosenfeld1968distance} to the inverted anonymization mask $(1 - \mathbf{M}_A)$ . This operation yields a topographical proximity map $\mathbf{D} \in \mathbb{R}^{H \times W}$, where each scalar value denotes the exact Euclidean or Manhattan distance to the nearest censored boundary.

We assign a priority score to each valid candidate coordinate $(x,y) \in \mathcal{V}$ by sampling the Distance Transform at the geometric centroid of the proposed patch: $\mathbf{D}(x + \frac{w}{2}, y + \frac{h}{2})$. The candidate set is then strictly sorted in ascending order of these centroid distances. This sorting operation introduces a computational bottleneck when evaluated along all valid coordinates, specially for very high-resolution images. However, thanks to the stride parameter $s$, the set of valid coordinates is reduced, alleviating heavy computation at a minor drop in semantically extracted information (more details in Section \ref{ssec:sp_yield}). 

\begin{figure}
    \centering
    \includegraphics[width=0.95\columnwidth]{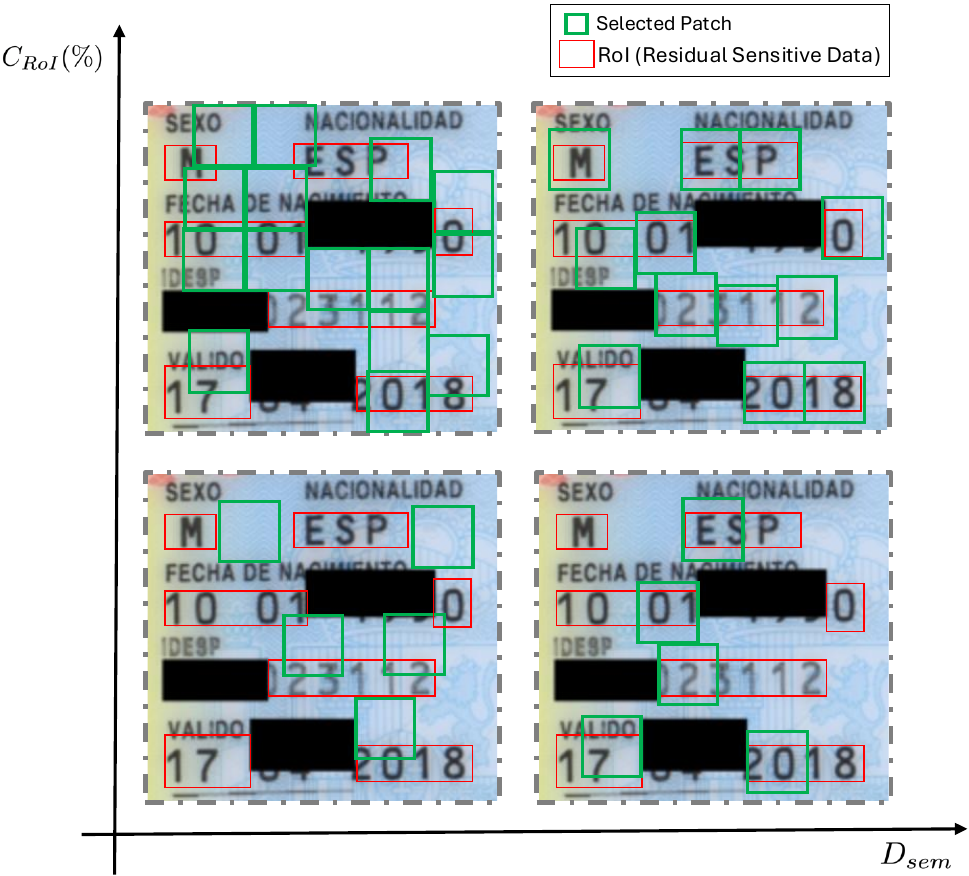}
    \caption{Visual examples of different patch extractions regarding ROI Coverage ($C_{RoI}$) and Semantic Density ($D_{sem}$) proposed metrics. It can be observed that, as both metrics increase, the patches contain more pixels from sensitive regions, which is related to the $D_{sem}$ metric, while also covering the most part of residual sensitive data, which relates to the $C_{RoI}$ metric.}
    \label{fig:metricexamples}
\end{figure}

\subsubsection{Greedy Spatial Non-Maximum Suppression (NMS)}
With the candidate list $\mathcal{C}$ sorted by boundary proximity, we must select the final patches while preventing spatial redundancy. We implement a Greedy Spatial Non-Maximum Suppression (NMS) utilizing a boolean occupancy grid $\mathbf{O} \in \{0, 1\}^{H \times W}$, initialized to zeros.

Iterating through the sorted list, a patch is accepted if and only if its entire spatial grid maps to zeros in $\mathbf{O}$ ($\sum \mathbf{O}[y:y+h, x:x+w] = 0$). Upon acceptance, this grid is updated to ones (occupied), effectively claiming the area. Because the list is ordered by distance, this greedy strategy forces the algorithm to ``crystallize" the patches around the perimeter of the anonymized regions first. Once the immediate border is saturated, subsequent valid patches naturally form consecutive outward layers.

Finally, for every selected coordinate, the corresponding visual data $\mathcal{E}_{\mathbf{I}}$ is extracted from the image $\mathbf{I}$, while the identical spatial footprint $\mathcal{E}_{GT}$ is extracted from the Ground Truth manipulation mask $\mathbf{M}_{GT}$. For \textit{real} \ac{id} images, $\mathbf{M}_{GT}$ acts as a tensor of zeros. This parallel extraction guarantees perfect spatial alignment between the image domain and the supervisory signal, optimizing the data pipeline for subsequent model training.

\begin{table}[!t]
    \centering
    \caption{Spatial Yield and Semantic Density Comparison on 250 Real Pseudo-Anonymized IDs for 128 $\times$ 128 patch size. Best result is highlighted in \textbf{bold}. Second best result is \underline{underlined}.}
    \label{tab:spatial_yield}
    \footnotesize
    \begin{tabular*}{\columnwidth}{@{\extracolsep{\fill}}lccc@{}}
        \toprule
        \textbf{Method} & 
        \textbf{\begin{tabular}{@{}c@{}}ROI Cov.\\(\%) $\uparrow$\end{tabular}} & 
        \textbf{\begin{tabular}{@{}c@{}}Sem.Dens.\\(px/patch)  $\uparrow$\end{tabular}} & 
        \textbf{\begin{tabular}{@{}c@{}}Time/Img.\\(s) $\downarrow$\end{tabular}} \\ 
        \midrule 
        Naive Grid \cite{fakeidet2} & 63.41 & 3,045 & \textbf{0.003} \\
        Kochkarev \textit{et al.} \cite{kochkarev2020data} & 52.76 & 3,435 & \underline{0.061} \\
        PatchMatch \cite{barnes2009patchmatch} & 45.13 & 3,765 & 0.077 \\
        \cmidrule{1-4}
        PACE ($s=1$)                & \underline{73.16} & \underline{3,804} &  6.127 \\ 
        PACE ($s=2$)                & \textbf{73.64} & \textbf{3,805} &  1.626 \\ 
        PACE ($s=8$)                & 73.15 & 3,753 &  0.135 \\
        \bottomrule
    \end{tabular*}
\end{table}

\subsection{PACE: Spatial Yield and Semantic Density Evaluation}
\label{ssec:sp_yield}

To ensure a rigorous comparative analysis under strict privacy constraints, all evaluated patch-extraction algorithms inherently enforce an absolute zero-overlap policy with the anonymization mask via $\mathcal{O}(1)$ integral image validation. We contrast PACE against topological adaptations of prominent state-of-the-art samplers. Specifically, the methodology by Kochkarev \textit{et al.} \cite{kochkarev2020data} is implemented as a probabilistic sampler weighted by the normalized inverse distance transform. Concurrently, the stochastic local search inspired by the random search phase of the PatchMatch algorithm \cite{barnes2009patchmatch} is also considered as a competing method for optimal patch selection. This unified framework isolates the geometric efficacy of each sampling strategy, evaluating their capacity to securely patch peri-sensitive regions, where the residual sensitive data lies.

We define two metadata-driven metrics to evaluate extraction efficacy: \textit{ROI Coverage} ($C_{RoI}$) and \textit{Semantic Density} ($D_{sem}$). $C_{RoI}$ measures the global percentage of critical forensic pixels successfully retrieved, whereas $D_{sem}$ evaluates the average amount of informative pixels packed into each patch, serving as a direct indicator of residual sensitive data coverage and purity. Let $\mathcal{R}$ denote the set of exposed \ac{roi} pixels, defined as the ground-truth \ac{id} bounding boxes excluding the anonymization mask. Given a set of $N$ valid extracted patches $\mathcal{E} = \{E_1, E_2, \dots, E_N\}$ extracted by an algorithm, the metrics are formalized as:
$$C_{RoI} (\%) = \frac{1}{|\mathcal{R}|} \sum_{n=1}^{N} |E_i \cap \mathcal{R}| \times 100, \quad D_{sem} = \frac{1}{N} \sum_{n=1}^{N} |E_i \cap \mathcal{R}|$$
where $|E_n \cap \mathcal{R}|$ represents the exact number of pristine, sensitive pixels captured by the $n$-th patch. 

Fig. \ref{fig:metricexamples} depicts these metrics visually. On the bottom left, we see a patch extraction where the semantic density ($D_{sem}$) and \ac{roi} Coverage ($C_{RoI}$) are low, as patches do not overlap with any of the \ac{roi} regions and no sensitive pixels are contained in the region the patch encompasses. The top left image shows the case of high $C_{RoI}$ and low $D_{sem}$, as patches cover the majority of the residual sensitive data, but for each patch, a fewer amount of sensitive pixels are contained. Conversely, in the bottom right image we see that patches contain a great amount of pixels belonging to residual sensitive data, but they do not cover them for the most part. Finally, the top right image showcases an ideal extraction, where patches are aligned with the residual sensitive data pixels, therefore producing purer patches, while also covering practically the whole area of residual sensitive information.

\begin{figure*}[!t]
    \centering
    \includegraphics[width=0.975\textwidth]{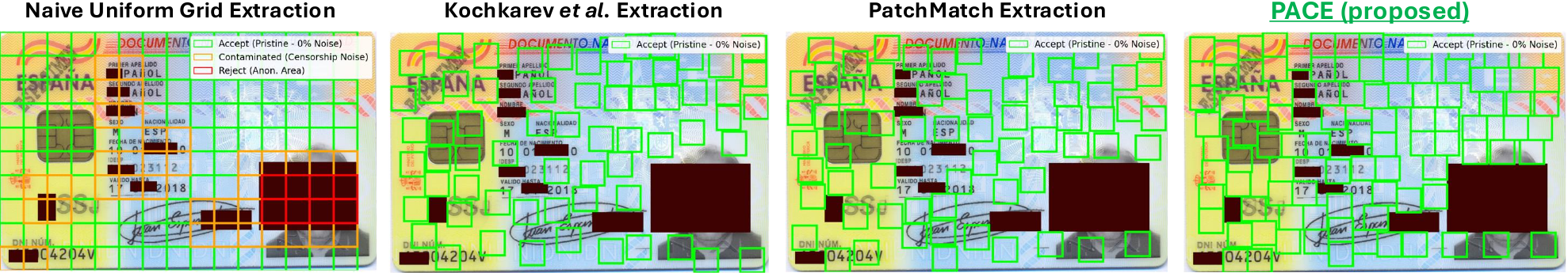}
    \caption{Comparison in terms of valid extracted 128$\times$128 patches (green), contaminated patches (orange) vs non-valid patches (red) in pseudo-anonymized configurations between the Naive Uniform Grid Method \cite{fakeidet2}, Kochkarev \textit{et al.} \cite{kochkarev2020data}, PatchMatch \cite{barnes2009patchmatch} and our proposed Pseudo-Anonymized Contextual patch Extraction method, PACE. It can be observed that the proposed PACE captures higher patch density in residual sensitive areas without compromising general coverage of the whole \ac{id} image compared with the competing methods thanks to a greedy strategy that prioritizes coordinates adjacent to redacted areas.}
    \label{fig:patch_extr_viz}
\end{figure*}

Table \ref{tab:spatial_yield} presents the evaluation across 250 real pseudo-anonymized IDs at 128$\times$128 patch resolution, benchmarking our approach against both rigid structural baselines and stochastic state-of-the-art methodologies. Evaluations were carried out in our server, with an Intel(R) Xeon(R) Gold 5420+ CPU using just one core. The limitations of a rigid uniform grid are evident: due to severe phase misalignment against irregular censorship boundaries, the naive grid captures only 63.41\% of the available \ac{roi}. The introduction of probabilistic sampling (Kochkarev \textit{et al.} \cite{patchforensics}) and stochastic perturbation (PatchMatch \cite{barnes2009patchmatch}) mitigates background dilution, yielding higher Semantic Densities (3,435 and 3,765 px/patch, respectively). However, this randomness causes severe structural fragmentation. By selectively skipping certain regions, these methods introduce significant discontinuities along the perimeter, reducing the \ac{roi} coverage below 53\%. By operating free of grid constraints while maintaining absolute deterministic packing, our proposed PACE dynamically contours the masks without leaving perimeter gaps. This strategic packing boosts \ac{roi} coverage to a leading 73.64\% and maximizes Semantic Density (3,805 pristine pixels/patch) at an optimal evaluation stride ($s=2$). Fig. \ref{fig:patch_extr_viz} depicts this phenomena visually, where it can be observed that our proposed patch greedy strategy extract patches that contour the redacted areas, better capturing the residual sensitive data. Conversely, the competing methods either capture contaminated areas (orange cells), or fail to properly occupy semantically dense regions where residual sensitive data is uncovered. We would also like to note that our proposed method performance degrade very slightly as the stride parameter $s$ goes up, showcasing its flexibility for practical applications.

\section{FakeIDet3-DB: Experimental Results}
\label{sec:eval_bchmrk}

FakeIDet3-DB pioneers the inclusion of digital attacks on real, government-issued \acp{id}. In this section we benchmark these attacks across both fake \ac{id} detection and localization tasks, utilizing standard metrics in the literature such as EER (\%) \cite{tapia2024first, tapia2025second} and pixel-level AUC (\%) \cite{trufor, difforensics, remtkd}, respectively. For this analysis, we select five state-of-the-art image forensic methods as baselines: TruFor \cite{trufor} and Re-MTKD \cite{remtkd} (denoted as Re-MTKD\textsubscript{AAAI25}) target general image forensics and perform both tasks. Additionally, we include its fine-tuned version---the DeepID challenge winner \cite{Korshunov_2025_ICCV}---denoted as Re-MTKD\textsubscript{ICCV25}, to assess how pre-trained domain data affects performance. Due to architectural specifics, SparseViT \cite{sparsevit} is evaluated exclusively on localization, while FakeIDet2 \cite{fakeidet2} serves strictly as a detection baseline given its original focus on physical \ac{id} attacks. Our primary objective is to evaluate how challenging the detection of FakeIDet3-DB's attacks is by state-of-the-art detectors.

Section \ref{sec:attack_diff_assess} evaluates the state-of-the-art detectors presented using the entire database (train and test sets) without anonymization, feeding the whole \ac{id} image. This allows a direct comparison to other public databases in the literature based on digital attacks. Section \ref{sec:loc_results} evaluates the detectors on our proposed patch-based benchmark, adhering to the privacy-aware protocol from \cite{fakeidet2}, using patches extracted from pseudo-anonymized \acp{id} via our proposed PACE patch extractor. All evaluations have been carried out using a single NVIDIA RTX 4090 GPU with weights strictly frozen.

\subsection{Attack Difficulty Assessment}
\label{sec:attack_diff_assess}

We evaluate the whole FakeIDet3-DB (i.e., no train/test split) to effectively demonstrate the quality and diversity of the proposed attacks. Given the wide range of attacks and post-processing methods, we divide the analysis into Classical and GenAI Attacks. Table \ref{tab:performance_classic_attacks} shows the performance on Classical Attacks, where we can see that the detectors struggle to detect and locate the proposed classical attacks. Regarding detection, we observe that both TruFor and Re-MTKD\textsubscript{ICCV25} perform quite on par overall (31.14\% \ac{eer} vs. 32.09\% \ac{eer}). Nevertheless, Re-MTKD\textsubscript{ICCV25} is better at detecting manipulations involving existing content (i.e., Copy-Move and Splicing) while TruFor excels detecting face morphing manipulations, which indicates that is more robust to face manipulations. While being optimized to detect physical \ac{id} attacks, FakeIDet2 performs quite poorly overall when detecting digital manipulations, which suggests that domain adaptation may not be a crucial factor for digital \ac{id} attacks detection. This phenomena is exacerbated for Re-MTKD\textsubscript{AAAI25}, where the performance is near random with 48.85\% \ac{eer}. 

\begin{table*}[t]
    \centering
    \caption{Performance on classical attacks. Best result is highlighted in \textbf{bold}. Second best result is \underline{underlined}.}
    \begin{tabular}{l ccccccc}
        \toprule
        & \multicolumn{2}{c}{\textbf{Copy-Move}} & \multicolumn{2}{c}{\textbf{Splicing}} & \multicolumn{2}{c}{\textbf{Face Morph.}} & \multirow{2}{*}{\textbf{Avg.}}\\
        \cmidrule(lr){2-3} \cmidrule(lr){4-5} \cmidrule(lr){6-7}
        \textbf{Model} & \textbf{Cheapfake} & \textbf{Refined} & \textbf{Cheapfake} & \textbf{Refined} & \textbf{Cheapfake} & \textbf{Refined} \\
        \midrule
        \multicolumn{7}{l}{\textbf{Detection (EER\% $\downarrow$)}} \\
        \midrule
        TruFor\textsubscript{\textcolor{gray}{[CVPR23]}} \cite{trufor} & \underline{32.80} &  \underline{43.21} &  \underline{19.60} &  \underline{33.65} & \textbf{22.43} &  \textbf{35.20} &  \textbf{31.14} \\
        Re-MTKD\textsubscript{\textcolor{gray}{[AAAI25]}} \cite{remtkd} & 50.04 & 52.21 & 43.60 & 48.41 & 50.49 & \underline{48.39} & 48.85 \\
        Re-MTKD\textsubscript{\textcolor{gray}{[ICCV25]}} \cite{Korshunov_2025_ICCV} &  \textbf{17.20} &  \textbf{40.00} & \textbf{10.11} &  \textbf{31.19} &  \underline{42.20} & 51.89 &  \underline{32.09} \\
        FakeIDet2\textsubscript{\textcolor{gray}{[InfFus26]}} \cite{fakeidet2} & 45.19 & 47.39 & 45.18 & 48.91 & 48.81 & 48.81 & 47.38 \\
        \midrule
        \multicolumn{7}{l}{\textbf{Localization (AUC\% $\uparrow$)}} \\
        \midrule
        TruFor\textsubscript{\textcolor{gray}{[CVPR23]}} \cite{trufor} & \textbf{96.35} & \textbf{88.75} &  \textbf{97.25} &  \textbf{89.19} & \underline{84.00} & \underline{75.58} & \textbf{88.52} \\
        Re-MTKD\textsubscript{\textcolor{gray}{[AAAI25]}} \cite{remtkd} & 71.10 & 68.39 & 81.81 & 68.39 &  \textbf{92.45} & \textbf{89.75} & 78.64 \\
        Re-MTKD\textsubscript{\textcolor{gray}{[ICCV25]}} \cite{Korshunov_2025_ICCV} &  \underline{93.32} &  \underline{81.63} &  \underline{95.39} &  \underline{81.12} & 76.35 & 61.93 &  \underline{81.62} \\
        SparseViT\textsubscript{\textcolor{gray}{[AAAI25]}} \cite{sparsevit} & 83.89 & 74.74 & 92.87 & 78.36 & 59.87 & 50.20 & 73.25 \\
        \bottomrule
    \end{tabular}
    \label{tab:performance_classic_attacks}
\end{table*}

Analyzing the localization task, TruFor outperforms the other detectors by an evident margin across most of the manipulation types with an average of 88.52\% AUC, which we believe is motivated by \textit{i)} the introduction of forensic priors, such as the Noiseprint++ module, which extracts residual forensic information from the original image, and \textit{ii)} the model was pre-trained with data containing copy-move and splicing attacks. Conversely, SparseViT yielded the lowest performance on classical attacks (73.25\% AUC). We attribute this to two main factors: the artifact-inducing downsampling of high-resolution \acp{id} to 512$\times$512, and the architecture's sparse-attention module. By restricting dense interactions between adjacent patch embeddings, sparse attention may dilute the weights in localized tampered regions, severely hindering the accurate localization of small manipulations.

\begin{table*}[t]
    \centering
    \caption{Performance on GenAI Attacks. Best result is highlighted in \textbf{bold}. Second best result is \underline{underlined}.}
    \setlength{\tabcolsep}{6pt}
    \begin{tabular}{l ccccccccc}
        \toprule
        & \multicolumn{2}{c}{\textbf{Text Inp.}} & \multicolumn{2}{c}{\textbf{Face Swap.}} & \multicolumn{2}{c}{\textbf{Face Morph.}} & \textbf{Cont. Remov.} & \textbf{Multi-Attack} & \multirow{2}{*}{\textbf{Avg.}}\\
        \cmidrule(lr){2-3} \cmidrule(lr){4-5} \cmidrule(lr){6-7} \cmidrule(lr){8-8} \cmidrule(lr){9-9}
        \textbf{Model} & \textbf{Cheapfake} & \textbf{Refined} & \textbf{Cheapfake} & \textbf{Refined} & \textbf{Cheapfake} & \textbf{Refined} & \textbf{Refined} & \textbf{Refined} \\
        \midrule
        \multicolumn{9}{l}{\textbf{Detection (EER\% $\downarrow$)}} \\
        \midrule
        TruFor\textsubscript{\textcolor{gray}{[CVPR23]}} \cite{trufor} & 40.40 & 46.00 & \underline{17.65} & \underline{15.07} & \underline{7.02} & \underline{17.96} &  \underline{50.00} & \underline{16.07} & \underline{26.27}  \\
        Re-MTKD\textsubscript{\textcolor{gray}{[AAAI25]}} \cite{remtkd} & \underline{37.64} & \underline{40.54} & 43.87 & 41.07 & 26.49 & 29.61 & 52.00 & 38.38 & 38.69  \\
        Re-MTKD\textsubscript{\textcolor{gray}{[ICCV25]}} \cite{Korshunov_2025_ICCV} &  \textbf{3.55} &  \textbf{6.67} & \textbf{5.09} & \textbf{4.26} & \textbf{4.02} & \textbf{15.46} & \underline{50.00} & \textbf{8.44} & \textbf{12.18} \\
        FakeIDet2\textsubscript{\textcolor{gray}{[InfFus26]}} \cite{fakeidet2} & 53.37 & 54.80 & 43.66 & 51.78 & 51.81 & 47.00 & \textbf{48.78} & 54.81 & 50.75  \\
        \midrule
        \multicolumn{9}{l}{\textbf{Localization (AUC\% $\uparrow$)}} \\
        \midrule
        TruFor\textsubscript{\textcolor{gray}{[CVPR23]}} \cite{trufor} & \textbf{93.57} & \textbf{86.02} & 81.22 & 77.67 & 84.48 & 75.04 & 50.20 & \underline{79.31} & 78.44  \\
        Re-MTKD\textsubscript{\textcolor{gray}{[AAAI25]}} \cite{remtkd} & 76.62 & 69.47 & \underline{89.70} & \textbf{92.17} & \underline{98.33} & \textbf{96.37} & \textbf{57.97} & \textbf{85.84} & \textbf{83.30} \\
        Re-MTKD\textsubscript{\textcolor{gray}{[ICCV25]}} \cite{Korshunov_2025_ICCV} & \underline{89.88} & \underline{71.56} & \textbf{98.29} & \underline{90.03} & 91.61 & \underline{86.35} & 51.32 & 68.68 & \underline{80.96} \\
        SparseViT\textsubscript{\textcolor{gray}{[AAAI25]}} \cite{sparsevit} & 84.07 & 69.70 & 60.55 & 72.93 & \textbf{99.91} & 83.21 & \underline{51.77} & 77.15 & 74.91 \\
        \bottomrule
    \end{tabular}%
    \label{tab:performance_genai_attacks}
\end{table*}

Regarding the detection of GenAI attacks, the results are presented in Table \ref{tab:performance_genai_attacks}. The first noticeable aspect is that Re-MTKD\textsubscript{ICCV25} detects GenAI attacks not only far better than Classical Attacks on average (12.18\% \ac{eer} vs. 32.09\% \ac{eer}), but also far better than the competing methods. For digital attacks with face and text manipulations, we observe that it achieves results under 10\% EER consistently, with the exception of refined face morphing (15.46\% EER) and content removal attacks (50\% EER). Regarding its pre-trained counterpart, Re-MTKD\textsubscript{AAAI25}, performance drops quite evidently across all types of attacks, regardless the refinement post-processing procedure. TruFor performs better detecting GenAI and Classical Attacks (26.27\% \ac{eer} vs. 31.14\% \ac{eer}), exhibiting mild generalization capabilities as it was not exposed to GenAI manipulations when pre-trained. However, for content removal attacks, we observe that no detector outperforms random guessing. 

Regarding localization performance, the detectors perform comparably on classical attacks, with TruFor leading the group. Intriguingly, Re-MTKD\textsubscript{AAAI25} outperforms its fine-tuned version to IDs by a small margin (83.30\% AUC vs. 80.96\% AUC), revealing a clear decorrelation between detection and localization performance. This discrepancy can be explained by examining the Re-MTKD architecture \cite{remtkd}, which decouples these tasks into two distinct heads: a classification head that leverages bottleneck features from the Cue-Net, and a localization head fed by the decoder. Because these heads are optimized using different objective functions, their respective weight updates are applied independently. Consequently, this separate optimization can cause the decoder to overfit to highly specific, dense localization cues present only in FantasyID, which fail to generalize to FakeIDet3-DB due to the more complex nature of the spatial supervisory signal. In contrast, the classification head extracts more generic, high-level features that remain applicable across domains, explaining why the model remains proficient at detection but fails at localization. TruFor's performance in GenAI attacks excels in Text Inpainting attacks (93.57\% AUC and 86.02\% AUC) with respect to the rest of the models, while struggling with face manipulations from generative models (from 75.04\% AUC to 84.48\% AUC). Furthermore, TruFor localization performance is worse in GenAI than Classic Attacks (78.44\% AUC vs. 88.52\% AUC), which is expected as its pre-training data encompassed only classic manipulations. Finally SparseViT remains the worst performing model in GenAI attacks (74.91\%), although it has the best performance in cheapfake face morphing attacks (99.91\% AUC), which can be considered the easiest attack due to the evident manipulation traces left. Additionally, we would like to note that our refinement post-processing produces significantly more challenging threats than naive replacement, degrading detection and localization performance compared to cheapfakes in 83.33\% and 91.67\% of cases, respectively.

\begin{table}[t]
    \centering
    \caption{Comparison of the proposed FakeIDet3-DB with FantasyID \cite{fantasyid} in terms of difficulty. Best result is highlighted in \textbf{bold}.}
    \footnotesize
    \setlength{\tabcolsep}{4pt}
    \begin{tabular}{lccc} 
        \toprule
        & & \makecell[c]{\textbf{TruFor}\cite{trufor} \\ \textcolor{gray}{\footnotesize CVPR23}} & \makecell[c]{\textbf{Re-MTKD}\cite{remtkd} \\ \textcolor{gray}{\footnotesize AAAI25}}  \\ 
        \cmidrule{3-4}
        \textbf{Task} & \textbf{Database} &  \\
        \midrule
        \multirow{2}{*}{\textbf{Det. (EER\% $\downarrow$)}} & FantasyID\cite{fantasyid} & \textbf{16.02} & \textbf{33.67} \\
        & FakeIDet3-DB & 32.45 & 40.28 \\
        \midrule
        \multirow{2}{*}{\textbf{Loc. (AUC\% $\uparrow$)}} & FantasyID\cite{fantasyid} & 82.38 & 75.01 \\
        & FakeIDet3-DB & \textbf{83.48} & \textbf{80.82} \\ 
        \bottomrule
    \end{tabular}
    \label{tab:comparison_difficulty_db}
\end{table}

Furthermore, for completeness, we compare the variability and quality of the digital attacks generated in our proposed FakeIDet3-DB with FantasyID \cite{fantasyid}, a recent public database in the field. This comparison is depicted in Table \ref{tab:comparison_difficulty_db} for both detection and localization tasks. For the localization task we leave out the real \acp{id}, therefore comparing strictly the fidelity on the digital attacks. For the detection task, we strictly consider our government-issued \acp{id} as the only real class, since FantasyID's ``real" \acp{id} are synthetically created in lab conditions. We would like to remark that we use the models' pre-trained weights frozen, therefore, they are not biased towards detecting any of the manipulations from any of the databases. Consequently, we leave out the Re-MTKD\textsubscript{ICCV25} as it was fine-tuned using FantasyID for the DeepID challenge. 

As can be seen in Table \ref{tab:comparison_difficulty_db}, it becomes evident that FakeIDet3-DB poses a significantly harder challenge for image-level detection. Specifically, the \ac{eer} of TruFor drastically increases from 16.02\% in FantasyID to 32.45\% in FakeIDet3-DB, doubling the error rate. A similar trend occurs for Re-MTKD, whose EER rises from 33.67\% to 40.28\%. 

Conversely, for the localization task, both detectors exhibit a slightly higher pixel-level AUC on FakeIDet3-DB (83.48\% and 80.82\% for TruFor and Re-MTKD, respectively) compared to FantasyID. This apparent paradox between poor global detection and competent local segmentation reveals a critical insight: while the manipulations in FakeIDet3-DB leave sufficient local traces to be relatively distinguished from the authentic background at a pixel level---which may be induced by the intricate patterns and engravings embedded only in real \acp{id}, being difficult to replicate by GenAI models---the overall image context perfectly camouflages these anomalies from a global classification perspective. Ultimately, these results demonstrate that FakeIDet3-DB acts as a demanding benchmark that effectively exposes the vulnerabilities of current forensics detectors under realistic, in-the-wild scenarios.

\begin{table}[!t] 
\caption{Dataset Split by Attack Modality, Models, and Refinement Level}
\label{tab:dataset_split}
\centering
\setlength{\tabcolsep}{4pt} 
\begin{tabular}{@{}llcc@{}}
\toprule
\multirow{2}{*}{\textbf{Modality}} & 
\multirow{2}{*}{\textbf{Models}} & 
\makecell{\textbf{Train} \\ \textit{(Cheapfake)}} & 
\makecell{\textbf{Test} \\ \textit{(Cheapfake} \\ \textit{\& Refined)}} \\
\midrule
\multirow{3}{*}{\textbf{\makecell[l]{GenAI Text \\ Attacks}}} & 
\makecell[l]{DiffSTE \cite{diffste}, RS-STE \cite{rsste}, \\ UDiffText \cite{udifftext}} & \textcolor{green}{\checkmark} & \textcolor{green}{\checkmark} \\
\addlinespace
& \textit{TextDiffuser 2 \cite{textdiff2}} & \textcolor{red}{\xmark} & \textcolor{green}{\checkmark} \\

\midrule 

\multirow{3}{*}{\textbf{\makecell[l]{GenAI Face \\ Swapping}}} & 
\makecell[l]{REFace \cite{reface}, \\ FaceDancer \cite{facedancer}} & \textcolor{green}{\checkmark} & \textcolor{green}{\checkmark} \\
\addlinespace
& \textit{InsightFace\footnotemark} & \textcolor{red}{\xmark} & \textcolor{green}{\checkmark} \\

\midrule

\multirow{2}{*}{\textbf{\makecell[l]{GenAI Face \\ Morphing}}} & 
FreeMorph \cite{freemorph} & \textcolor{green}{\checkmark} & \textcolor{green}{\checkmark} \\
\addlinespace
& \textit{DiffMorpher \cite{diffmorph}} & \textcolor{red}{\xmark} & \textcolor{green}{\checkmark} \\

\midrule

\textbf{Classical} & All Classical Attacks & \textcolor{green}{\checkmark} & \textcolor{green}{\checkmark} \\

\bottomrule
\end{tabular}
\end{table}

\footnotetext{\href{https://github.com/deepinsight/insightface}{https://github.com/deepinsight/insightface}}

\subsection{Public Benchmark}
\label{sec:loc_results}

\begin{table*}[t]
    \centering
    \caption{Proposed patch-based benchmark: Performance on both localization and detection tasks in the pseudo-anonymized scenario. Best result is highlighted in \textbf{bold}. Second best is \underline{underlined}.}
    \setlength{\tabcolsep}{4pt}
    \resizebox{\textwidth}{!}{%
        \begin{tabular}{lccccccccc} 
            \toprule
            & \multicolumn{3}{c}{\textbf{Classical Attacks}} & \multicolumn{5}{c}{\textbf{GenAI Attacks}} & \\ 
            \cmidrule(lr){2-4} \cmidrule(lr){5-9}
            \textbf{Model} & \textbf{Copy-Move} & \textbf{Splicing} & \textbf{Face Morph.} & \textbf{Text Inp.} & \textbf{Face Swap.} & \textbf{Face Morph.} & \textbf{Cont. Remov.} & \textbf{Multi-Attack} & \textbf{All} \\
            \midrule
            \multicolumn{10}{l}{\textbf{Detection (EER\% $\downarrow$)}} \\
            \midrule
            \multicolumn{10}{c}{\textit{Patch Size:} 128$\times$128} \\
            \midrule 
            TruFor\textsubscript{\textcolor{gray}{[CVPR23]}} \cite{trufor} & 47.83 & 45.65 & 54.35 & 40.87 & 54.35 & \underline{34.78} & \underline{47.83} & 43.48 & 47.83 \\              
            Re-MTKD\textsubscript{\textcolor{gray}{[AAAI25]}} \cite{remtkd} & 47.83 & \textbf{34.78} & 41.31 & \underline{35.32} & \underline{36.52} & 42.39 & \underline{47.83} & \textbf{26.09} & \underline{39.13} \\
            Re-MTKD\textsubscript{\textcolor{gray}{[ICCV25]}} \cite{Korshunov_2025_ICCV} & \textbf{34.78} & \textbf{34.78} & \textbf{39.95} & \textbf{31.49} & \textbf{29.56} & \textbf{23.91} & \textbf{34.78} & \textbf{26.09} & \textbf{34.78} \\
            FakeIDet2\textsubscript{\textcolor{gray}{[InfFus26]}} \cite{fakeidet2} & \underline{45.65} & 47.83 & \underline{41.30} & 44.57 & 42.61 & 46.74 & \underline{47.83} & 47.83 & 47.48 \\
            \midrule
            \multicolumn{10}{c}{\textit{Patch Size:} 64$\times$64} \\
            \midrule 
            TruFor\textsubscript{\textcolor{gray}{[CVPR23]}} \cite{trufor} & \textbf{43.48} & \underline{47.83} & \textbf{47.83} & 51.62 & 51.30 & \underline{50.00} & \underline{47.83} & 47.83 & 52.17 \\ 
            Re-MTKD\textsubscript{\textcolor{gray}{[AAAI25]}} \cite{remtkd} & \underline{52.17} & \textbf{45.65} & \underline{54.34} & \underline{37.49} & \underline{46.09} & 58.69 & \underline{47.83} & \textbf{30.43} & \textbf{47.48} \\
            Re-MTKD\textsubscript{\textcolor{gray}{[ICCV25]}} \cite{Korshunov_2025_ICCV} & 54.34 & 52.18 & 58.70 & 50.54 & 54.78 & 57.61 & 52.17 & 39.13 & 52.17 \\
            FakeIDet2\textsubscript{\textcolor{gray}{[InfFus26]}} \cite{fakeidet2} & 56.62 & 56.62 & 56.52 & \textbf{30.43} & \textbf{43.47} & \textbf{47.82} & \textbf{43.48} & \textbf{30.43} & \textbf{47.48} \\
            \midrule
            \multicolumn{10}{l}{\textbf{Localization (AUC\% $\uparrow$)}} \\
            \midrule
            \multicolumn{10}{c}{\textit{Patch Size:} 128$\times$128} \\
            \midrule TruFor\textsubscript{\textcolor{gray}{[CVPR23]}} \cite{trufor} & \textbf{73.27} & \textbf{80.37} & 64.28 & \underline{62.82} & 59.69 & \textbf{63.69} & \underline{49.37} & \underline{48.17} & \textbf{62.71} \\ 
            Re-MTKD\textsubscript{\textcolor{gray}{[AAAI25]}} \cite{remtkd} & \underline{71.79} & 68.85 & \underline{68.10} & 62.02 & \textbf{69.19} & \underline{62.13} & 44.27 & \textbf{50.43} & \underline{62.09} \\
            Re-MTKD\textsubscript{\textcolor{gray}{[ICCV25]}} \cite{Korshunov_2025_ICCV} & 69.13 & \underline{72.12} & 48.23 & \textbf{68.62} & 51.43 & 48.12 & \textbf{52.17} & 34.78 & \underline{62.09} \\
            SparseViT\textsubscript{\textcolor{gray}{[AAAI25]}} \cite{sparsevit} & 69.64 & 41.43 & \textbf{68.85} & 47.95 & 43.99 & 45.39 & 45.59 & 45.27 & 51.03 \\
            \midrule
            \multicolumn{10}{c}{\textit{Patch Size:} 64$\times$64} \\
            \midrule 
            TruFor\textsubscript{\textcolor{gray}{[CVPR23]}} \cite{trufor} & \underline{54.63} & \underline{60.49} & \textbf{50.63} & \underline{43.18} & \textbf{51.81} & \textbf{56.67} & \textbf{53.52} & 38.67 & \textbf{51.18} \\ 
            Re-MTKD\textsubscript{\textcolor{gray}{[AAAI25]}} \cite{remtkd} & 39.87 & 53.31 & 38.29 & 35.79 & \underline{46.05} & \underline{46.25} & 43.09 & 32.20 & 41.85 \\
            Re-MTKD\textsubscript{\textcolor{gray}{[ICCV25]}} \cite{Korshunov_2025_ICCV} & \textbf{61.25} & \textbf{63.30} & 25.12 & \textbf{58.95} & 43.70 & 44.09 & 35.94 & \textbf{56.42} & \underline{48.69} \\
            SparseViT\textsubscript{\textcolor{gray}{[AAAI25]}} \cite{sparsevit} & 51.14 & 54.01 & \underline{40.08} & 42.79 & 38.61 & 36.77 & \underline{45.74} & \underline{42.20} & 43.91 \\
            \bottomrule
        \end{tabular}
    }
    \label{tab:benchmark_results}
\end{table*}

Finally, in this section we describe the details of our proposed benchmark, which is publicly available to the research community in order to advance in this challenging field. This benchmark shifts from full-image to patch-based \ac{id} processing (128$\times$128 and 64$\times$64 patch sizes) to safeguard sensitive data, as described throughout the article and our proposed PACE patch extractor (Sec.~\ref{sec:patch_extraction}). Following \cite{fakeidet2}, detection treats a single \ac{id} as an unordered collection of independent patches, outputting a global authenticity score. For localization, we evaluate patch-wise binary masks under strict privacy constraints, requiring models to generate spatial predictions exclusively for non-redacted inputs. As most baseline detectors were trained on full images, we applied here simple adaptations: detectors process individual patches to output local masks, and patch-level detection scores are averaged via mean fusion to compute the final \ac{id}-level score \cite{fakeidet}.

To simulate real-world conditions where detectors face unseen threats, we carefully partition our dataset to restrict the availability of both attack types and \ac{id} templates during training. Following the FakeIDet2-db protocol \cite{fakeidet2}, the first and second versions of the Spanish \ac{id} are reserved for testing, leaving the third exclusively for training. Furthermore, as detailed in Table \ref{tab:dataset_split}, the training set is strictly limited to classical attacks and cheapfake manipulations from a subset of GenAI models. Highly capable models and refined post-processing techniques are held out for the test set. This allows us to rigorously evaluate whether models merely overfit known forensic traces or appropriately generalize to unseen GenAI models and novel refinement procedures. Full details are available in our GitHub.

Table \ref{tab:benchmark_results} demonstrates that the restricted patch-based evaluation severely degrades detector capabilities, exposing their heavy reliance on global image context and macro-structural artifacts. In detection, error rates suffer a near-total collapse. At 128$\times$128 patch size, Re-MTKD\textsubscript{ICCV25} achieves the best performance (34.78\% \ac{eer}), yet remains unacceptably high for real scenarios. At 64$\times$64 patch size, detectors essentially revert to random guessing ($\ge$50\% \ac{eer}). Notably, FakeIDet2---despite its patch-based design---fails against digital manipulations (47.48\% \ac{eer}), proving digital forensic traces are vastly subtler than the physical forgeries it was optimized for. These results expose that naive mean-score fusion is insufficient to aggregate localized predictions as shown in \cite{fakeidet2} and more sophisticated strategies are needed for patch-based digital attacks detection.

The task of localization suffers a similar fate. At 128$\times$128 patch size, TruFor and Re-MTKD marginally segment Classical attacks (e.g., TruFor achieves 80.37\% AUC on Splicing), likely capturing abrupt high-frequency noise discrepancies within the patch. However, GenAI manipulations bypass detection entirely. At 64$\times$64 patch size, localization drops near 50\% AUC across almost all detectors, neutralizing their spatial predictive capabilities. This trend reflects an inherent issue in the baseline architectures when applied to patch-based localization, as their receptive fields---accustomed to high-resolution global semantics ($\ge$512$\times$512)---are limited by the restrictive semantic content embedded in individual patches.

\section{Conclusion}
\label{sec:conclusion}

In this article we presented FakeIDet3-DB, the first database featuring digital attacks on 250 images of real, government-issued \acp{id}. Ensuring a comprehensive set of classical and GenAI attacks, we presented a novel post-processing pipeline generating diverse digital attacks, bridging the gap between easy-to-spot \textit{cheapfakes} and highly realistic \textit{refined} forgeries. Evaluating state-of-the-art methods on full, non-anonymized \acp{id}, FakeIDet3-DB poses a harder detection challenge compared to existing databases like FantasyID~\cite{fantasyid} (32.45\% vs. 16.02\% \ac{eer}), while remaining competitive in localization (83.48\% vs. 82.38\% AUC). Our refined post-processing pipeline creates highly deceptive attacks, as in 83.33\% and 91.67\% of evaluated cases, refined attacks yielded worse detection and localization results than their cheapfake counterparts.

Because unrestricted image distribution violates data privacy regulations, we adopted a privacy-aware framework \cite{fakeidet2}, applying strategic anonymization via content redaction to prevent \ac{pii} leakage (i.e., pseudo-anonymization). Subsequently, we proposed PACE, a geometrically constrained patch extraction algorithm that prioritizes spatial coordinates adjacent to redacted regions. We demonstrated this greedy strategy achieves an optimal balance between residual sensitive information coverage and semantic density with low computational overhead, making it suitable for real-world scenarios.

Finally, we established a privacy-compliant benchmark leveraging 5.2M patches extracted from over 6.4K pseudo-anonymized real/fake \ac{id} images. Evaluating state-of-the-art detectors (Re-MTKD \cite{remtkd}, TruFor \cite{trufor}, SparseViT \cite{sparsevit}, FakeIDet2 \cite{fakeidet2}), we observed considerable performance degradation compared to full-image evaluations. We primarily attribute this drop to the baselines' receptive fields, optimized for higher resolutions. Moreover, the naive fusion strategies employed to aggregate patch scores further hinder performance.

Future work will explore the development of architectures specifically designed to detect and localize digital attacks while strictly complying with privacy regulations. We hope the research community embraces patch-based frameworks not as an impediment, but as a secure, alternative paradigm enabling vital cooperation between AI researchers and data holders.

\section*{Acknowledgment}
This project has been supported by PowerAI+ (SI4/PJI/2024- 00062 Comunidad de Madrid and UAM), Cátedra ENIA UAM-Veridas en IA Responsable (NextGenerationEU PRTR TSI-100927-2023-2), and TRUST-ID (PID2025-173396OB-I00 MICIU/AEI and the EU).

\bibliographystyle{IEEEtran}
\bibliography{bibtex/bib/IEEEabrv,bibtex/bib/refs, bibtex/bib/IEEEexample}

\end{document}